\documentclass[conference]{IEEEtran}
\IEEEoverridecommandlockouts

\usepackage{cite}
\usepackage{amsmath,amssymb,amsfonts}
\usepackage{algorithmic}
\usepackage{graphicx}
\usepackage{textcomp}
\usepackage{xcolor}
\usepackage{epstopdf}
\usepackage{booktabs}
\usepackage{tabularx}
\usepackage{array}
\usepackage{makecell}
\usepackage{multirow}
\usepackage{multicol}
\usepackage{tikz, pgfplots}
\usepackage{url}

\definecolor{Seaborn1}{HTML}{1f77b4}
\definecolor{Seaborn2}{HTML}{ff7f0e}
\definecolor{Seaborn3}{HTML}{2ca02c}
\definecolor{Seaborn4}{HTML}{d62728}

\def\BibTeX{{\rm B\kern-.05em{\sc i\kern-.025em b}\kern-.08em
    T\kern-.1667em\lower.7ex\hbox{E}\kern-.125emX}}

\begin{document}

\title{Gaze Estimation for Human-Robot Interaction: Analysis Using the NICO Platform
\thanks{This work was funded by the EU NextGenerationEU through the Recovery and Resilience Plan for Slovakia under Project No.09I02-03-V01-00029. The work presented in this paper was carried out in the framework of the TERAIS project, a Horizon-Widera-2021 program of the European Union under the Grant agreement number 101079338. This work was also supported by the Slovak Grant Agency for Science (VEGA), project no.~1/0373/23.}
}

\author{
\IEEEauthorblockN{1\textsuperscript{st} Matej Palider}
\IEEEauthorblockA{\textit{Faculty of Mathematics, Physics} \\
\textit{and Informatics} \\
\textit{Comenius University} \\
Bratislava, Slovakia}
\and
\IEEEauthorblockN{2\textsuperscript{nd} Omar Eldardeer}
\IEEEauthorblockA{\textit{CONTACT Unit} \\
\textit{Italian Institute of Technology} \\
Genoa, Italy \\
0000-0002-6029-3138}
\and
\IEEEauthorblockN{3\textsuperscript{rd} Viktor Kocur}
\IEEEauthorblockA{\textit{Faculty of Mathematics, Physics} \\
\textit{and Informatics} \\
\textit{Comenius University} \\
Bratislava, Slovakia \\
viktor.kocur@fmph.uniba.sk \\
0000-0001-8752-2685}
}

\maketitle

\begin{abstract}
This paper evaluates the current gaze estimation methods within a human-robot interaction (HRI) context of a shared workspace scenario. We introduce a new, annotated dataset collected with the NICO robotic platform. We evaluate four state-of-the-art gaze estimation models. The evaluation shows that the angular errors are close to those reported on general-purpose benchmarks. However, when expressed in terms of distance in the shared workspace the best median error is 14.57~cm, quantifying the practical limitations of current methods. We conclude by discussing these limitations and offering recommendations on how to best integrate gaze estimation as a modality in HRI systems.
\end{abstract}

\begin{IEEEkeywords}
HRI, gaze estimation, deep learning, dataset
\end{IEEEkeywords}

\section{Introduction}

Human gaze estimation is incorporated into many different human-robot interaction (HRI) applications \cite{admoni2017social}. This is mainly because of the importance of the gaze as a social non-verbal cue for interaction \cite{george2008facing}. It drives many different social cognitive mechanisms (such as joint attention, intention prediction, and task coordination) and provides an explainable behavior for others \cite{mundy2007individual}. The ability to perceive and understand the social cues affects the effectiveness and efficiency of the whole interaction experience. Gaze understanding and following is one of the earliest behavior mechanisms developed by infants to engage in different social communication scenarios \cite{beier2012infants}.

Therefore, achieving high accuracy in gaze estimation is a key enabler of seamless human-robot interaction. Despite the significant progress of gaze estimation methodologies, these methods have not yet been fully evaluated in real human-robot interaction scenarios.

In this paper we present an applied evaluation of the latest gaze estimation methods in a standard HRI scenario, specifically when the human and the robot are engaged in a shared workspace (e.g., table surface). With this paper, our contributions are: an embodiment for gaze estimation methods using the NICO robotic platform~\cite{kerzel2017nico}, a dataset for HRI gaze and an evaluation of the latest state-of-the-art off-the-shelf gaze estimation methods~\cite{cheng2022gaze,abdelrahman2023l2cs,ververas20243dgazenet,vuillecard20253dgaze}. We make our dataset and evaluation code available.\footnote{The dataset is available at \url{https://doi.org/10.5281/zenodo.23059771}\\ The evaluation code is available at \url{https://github.com/kocurvik/nico_gaze}}

The NICO platform was chosen for this study for its accessibility to the research community. As an open-source humanoid platform, NICO~\cite{kerzel2017nico} provides freely available mechanical designs, electronics schematics, and control software, which lowers the barrier to entry for HRI research compared to proprietary commercial platforms. Its relatively low cost and modular construction make it well suited not only for research but also for educational use, allowing institutions with limited budgets to conduct hands-on studies in humanoid robotics, human-robot interaction, and developmental robotics.

Our analysis reveals that the accuracy of the evaluated gaze estimation methods in our scenario is similar to that reported on general-purpose gaze estimation datasets~\cite{kellnhofer2019gaze360} with the best performing methods achieving mean errors slightly below 10$^\circ$. However, when evaluated in terms of distance error in the shared workspace the best performing method~\cite{ververas20243dgazenet} achieves a median error of 14.57~cm, thus providing only a rough estimate of the point where the human directs their gaze. This limitation should be considered when gaze estimation is included as a modality in HRI tasks.

\begin{figure*}[htbp]
    \centering
    \begin{tabular}{ccc}
    \includegraphics[height=0.14\textheight]{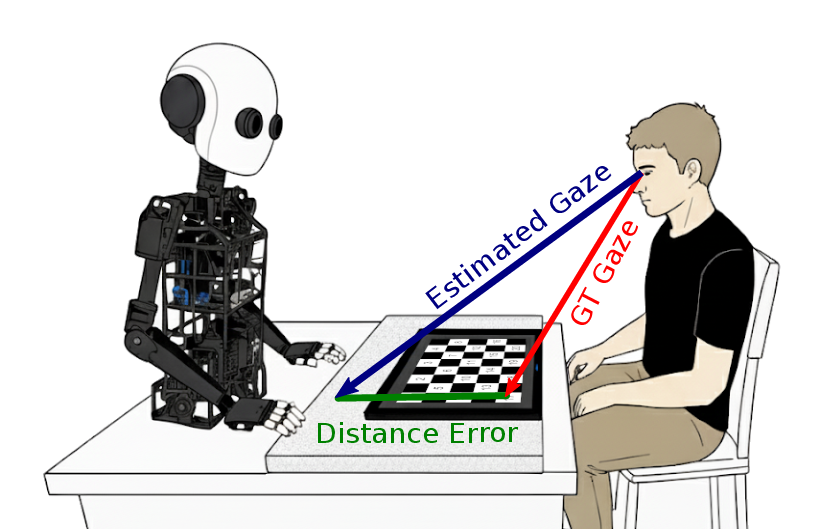} &
    \includegraphics[height=0.14\textheight]{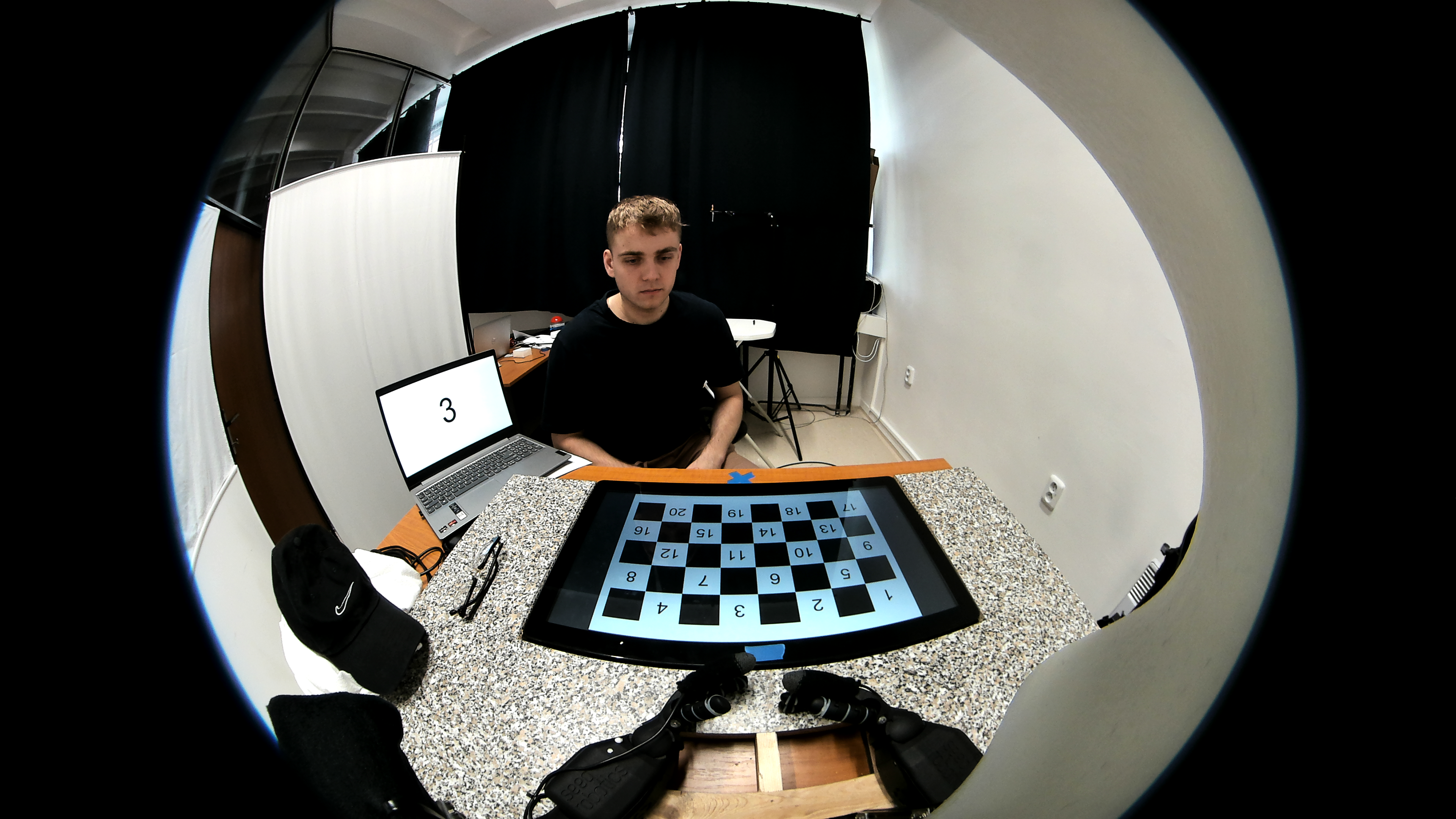} &
    \includegraphics[height=0.14\textheight]{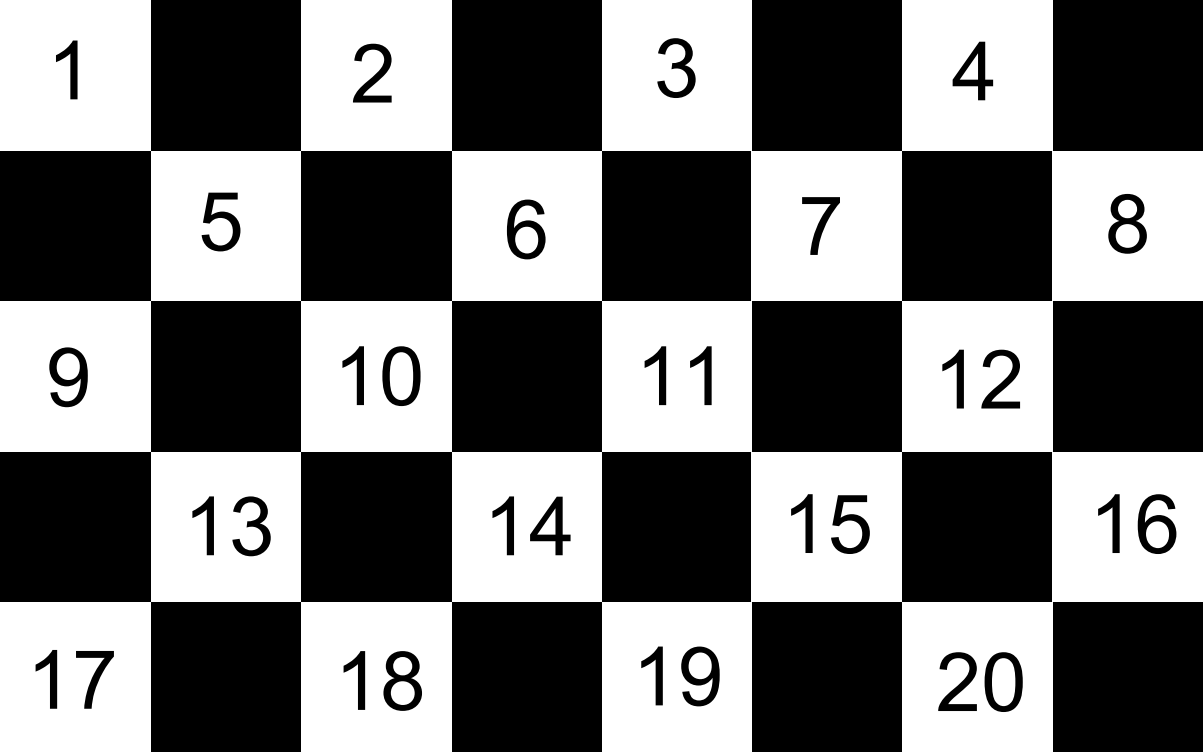}
    \\
    a) & b) & c)
    \end{tabular}
    \caption{a) The experimental setup. A human participant sits in front of the robot. A display is embedded within the shared workspace of the robot and the participant. In our study we consider the distance between the estimated and ground truth gaze point in the shared workspace. b) View from the robot's left camera. c) The grid shown on the display.}
    \label{fig:setup}
\end{figure*}

\section{Related Work}

\subsection{3D Gaze Estimation}

3D gaze estimation attempts to estimate the 3D direction of human gaze from images. As input, cropped images of eyes~\cite{biswas2021appearance}, faces~\cite{cheng2022gaze,abdelrahman2023l2cs,ververas20243dgazenet} or their combination~\cite{krafka2016eye} can be used. Some methods also focus on image sequences from videos to perform dynamic gaze estimation~\cite{vuillecard20253dgaze}. Most recent works are based on deep learning utilizing convolutional neural networks~\cite{abdelrahman2023l2cs,ververas20243dgazenet}, transformers~\cite{vuillecard20253dgaze}, or combined architectures~\cite{cheng2022gaze}. Typically, the neural networks directly regress sparse representations of gaze direction. However, the full 3D eyeball geometry can also be modeled to improve accuracy~\cite{ververas20243dgazenet}.

\subsection{Datasets}

Training deep neural networks generally requires sufficient training data. Several gaze estimation datasets have been published in recent years. Some datasets~\cite{zhang2017mpiigaze} feature only limited distribution of gaze directions, include few human subjects, or are constrained to specific environments or applications. Gaze360~\cite{kellnhofer2019gaze360} aims to remedy these issues by collecting a dataset with a wide distribution of gaze directions featuring 238 human subjects in various environments. ETH-XGaze~\cite{zhang2020ethxgaze} focuses on variations in terms of pose and illumination. 

To obtain the ground truth, setups with markers or projection systems and calibrated cameras or camera rigs are commonly used. These hardware requirements make the collection of large datasets very impractical. Some methods therefore aim to utilize datasets without annotated gaze. In~\cite{kothari2021weakly} the authors curated a large collection of people looking at each other to provide a weak supervision signal to train gaze estimation networks. In a similar fashion 2D gaze information~\cite{vuillecard20253dgaze}, or multi-view consistency~\cite{ververas20243dgazenet} can be used to provide the training signal. Lack of real training data can also be supplemented by utilizing synthetically generated data~\cite{ververas20243dgazenet,herashchenko2023appearance}.

\section{Gaze Estimation from Robot's Perspective}
\label{sec:method}

The goal of this study is to evaluate the accuracy and usefulness of off-the-shelf gaze estimation methods in a standard HRI setup. We consider a setup where a human interacts with a humanoid robot in a shared workspace with a dominant plane (e.g., a table). We implement a system which uses stereo-vision, face detection and gaze estimation methods to estimate the gaze point in the plane. In this section, we first provide details on our hardware setup and then introduce various parts of our gaze estimation pipeline.

\subsection{Hardware Setup}

In this paper we consider an HRI scenario with a humanoid robot. We use the NICO robotic platform \cite{kerzel2017nico} equipped with two See3CAM CU135 wide field-of-view cameras inside the robot's head. The robot's torso is placed on a table with a built-in display in the shared workspace. The display is used to perform calibration of the robot camera poses. The humans interacting with the robot sit opposite the robot and are directed to focus their gaze on individual squares on a grid displayed on the display. The setup is shown in Fig.~\ref{fig:setup}.

\begin{figure*}[htbp]
    \centering
    \includegraphics[height=0.125\textheight]{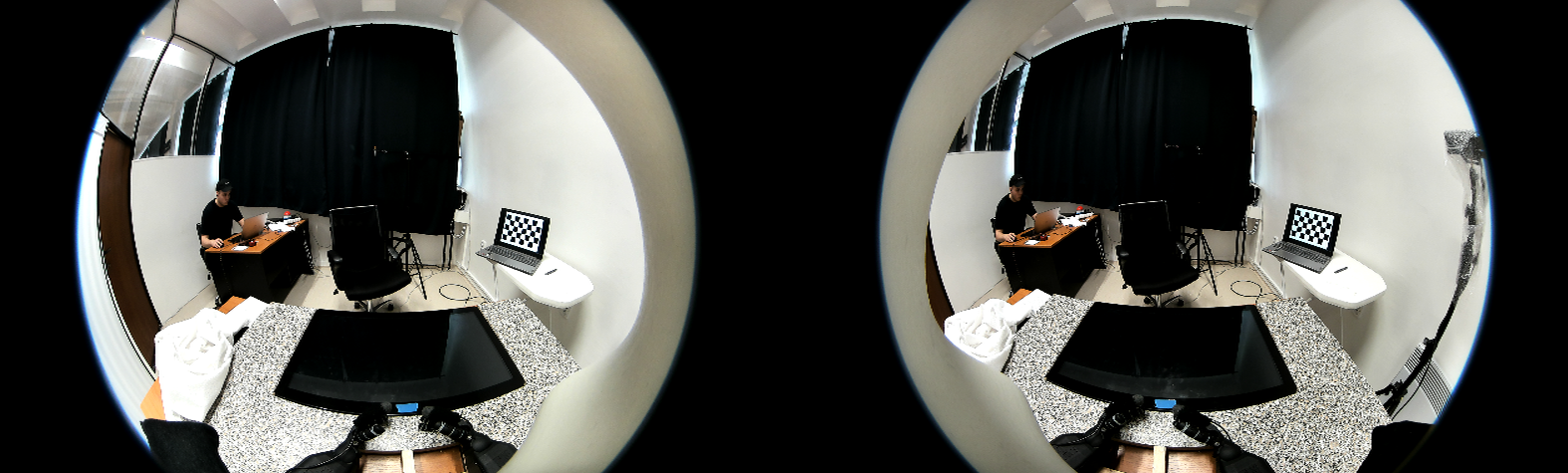} \hfill \includegraphics[height=0.125\textheight]{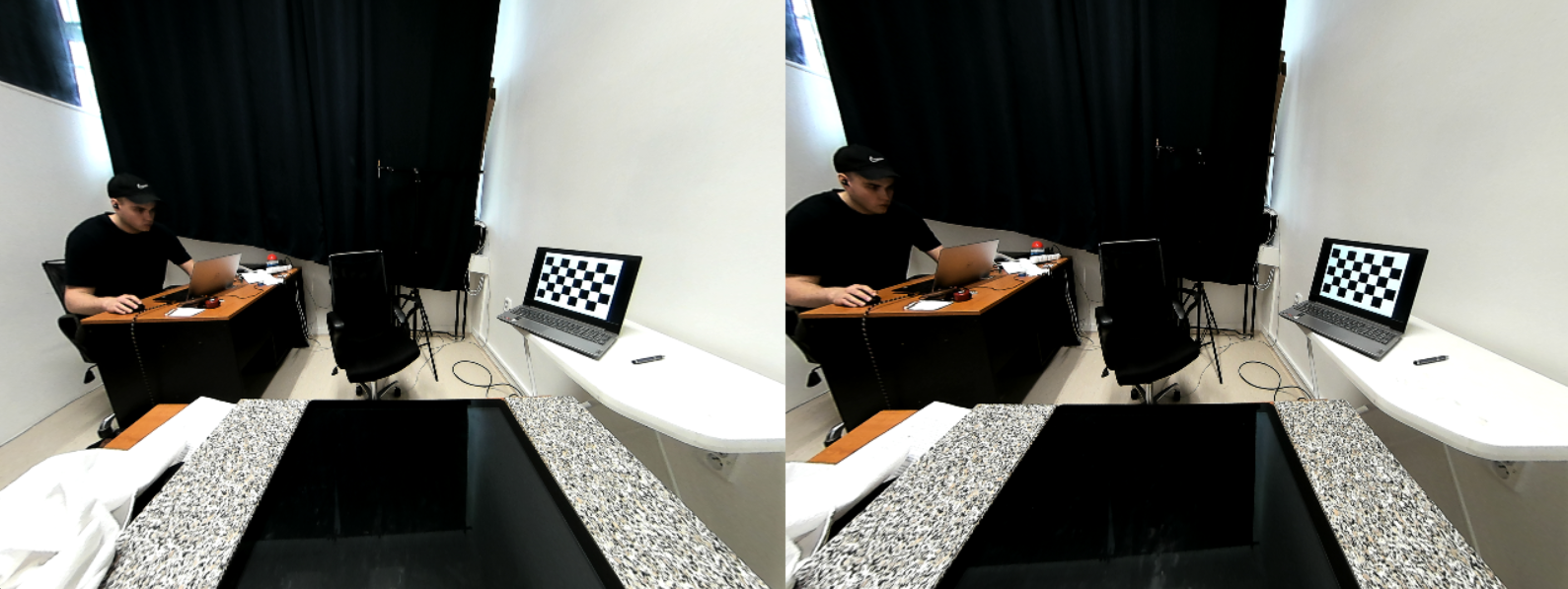}
    \caption{Example of images used for stereo-calibration. \textbf{Left:} Wide field-of-view cameras are significantly distorted. \textbf{Right:} With known camera intrinsics it is possible to rectify the images.}
    \label{fig:calib}
\end{figure*}

\subsection{Camera Calibration}
\label{sec:calib}

To infer the correct geometry using the robot's cameras, we perform camera calibration with a checkerboard pattern using the omnidirectional camera model implemented in OpenCV. We use images of the calibration pattern captured simultaneously by both cameras to also estimate their relative poses. The knowledge of the camera intrinsics allows us to rectify the images by removing the distortion caused by their lenses (see Fig.~\ref{fig:calib}). Since the relative poses of the cameras are also known, it is possible to triangulate points seen by both cameras to obtain a point in 3D space.

In our setup the humans are looking at a display embedded within the shared workspace in front of the robot. A checkerboard grid with the white squares numbered is shown on the display (see Fig.~\ref{fig:setup}). To evaluate the accuracy of the estimated gaze we need to know the transformation between two coordinate frames: $\mathcal{C}_c$, which is the coordinate frame of the robot's left camera, and $\mathcal{C}_\pi$, which is the coordinate frame associated with the plane of the display. In $\mathcal{C}_\pi$, the coordinates of the checkerboard corners are $\vec{Y}_{i, j} = (si, sj, 0)$, where $i$ and $j$ denote the row and the column of the grid respectively and $s$ denotes the metric square size (width and height). $\mathcal{C}_\pi$ is thus defined such that the shared workspace lies in the plane $\pi$ defined by $Z = 0$.

To detect the plane position with respect to $\mathcal{C}_c$, we first detect the checkerboard corners in the image using the OpenCV checkerboard detector. We establish 2D-3D correspondences between the detected 2D corners and 3D points $\vec{Y}_{i, j}$. Using these 2D-3D correspondences and \cite{collins2014infinitesimal} we estimate the transformation between $\mathcal{C}_c$ and $\mathcal{C}_\pi$ defined by the rotation $R_{c \rightarrow\pi}$ and translation $\vec{t}_{c\rightarrow\pi}$.

\begin{table*}[htbp]
    \centering
    \caption{The results of the evaluated methods for the full dataset.}
    \begin{tabular}{cccccc}
    \toprule
    \multirow{2}{*}{Method} & \multirow{2}{*}{\makecell{Mean Angular \\ Error ($^\circ$)}} & \multirow{2}{*}{\makecell{Median \\ Distance (cm)}} & \multicolumn{3}{c}{Distance Precision (\%)} \\
    & & & @10cm & @20cm & @50cm \\ \midrule
    GazeTR~\cite{cheng2022gaze} & 14.69 & 20.21 & 21.27 & 48.57 & 89.84 \\
    L2CS~\cite{abdelrahman2023l2cs} & \phantom{1}\textbf{9.65} & 15.63 & 33.02 & \textbf{66.03} & 90.48 \\
    3DGazeNet~\cite{ververas20243dgazenet} & \phantom{1}9.92 & \textbf{14.57} & \textbf{33.65} & 61.27 & 83.17 \\
    Gaze3D~\cite{vuillecard20253dgaze} & 11.97 & 19.31 & 20.95 & 54.29 & \textbf{96.51} \\
    \bottomrule
    \end{tabular}
    \label{tab:results}
\end{table*}

\subsection{Gaze Estimation}
\label{sec:method_gaze}

To estimate human gaze we test several off-the-shelf gaze estimation methods \cite{cheng2022gaze,abdelrahman2023l2cs,ververas20243dgazenet,vuillecard20253dgaze}. All of these methods use deep neural networks to estimate human gaze direction. As outputs, these methods either provide a normalized 3D vector expressed in Cartesian coordinates or the yaw and pitch of the gaze, which can be converted to a normalized 3D vector. More details regarding the implementation of the individual methods are provided in Sec.~\ref{sec:eval_nets}.

To estimate the gaze rays in $\mathcal{C}_c$ we first detect the bounding boxes of faces in images from both of the robot's cameras taken simultaneously. Using the known stereo-calibration, we can triangulate the center points of the detected bounding boxes. We denote this point as $\vec{x}$. Then we obtain the normalized gaze direction $\vec{d}$ from a gaze estimation network. The ray of the gaze in $\mathcal{C}_c$ is then $\{ \vec{x} + \lambda \vec{d}~|~0 \leq \lambda\}$. Using the transformation between $\mathcal{C}_c$ and $\mathcal{C}_\pi$ we can transform the gaze ray into $\mathcal{C}_\pi$ denoting it as $\{\vec{x}' + \alpha \vec{d}'~|~0 \leq \alpha\}$, where $\vec{x}' = R_{c \rightarrow\pi} \vec{x} + \vec{t}_{c \rightarrow\pi}$ and $\vec{d}' = R_{c \rightarrow\pi} \vec{d}$. The point of intersection with the plane $\pi$ in $\mathcal{C}_\pi$ is then found by solving for $\alpha = - x'_3 / d'_3$, where $x'_3$ and $d'_3$ denote the third coordinate of vectors $\vec{x}'$ and $\vec{d}'$ respectively.

\begin{figure}[htbp]
    \centering
    \begin{tikzpicture}
    \begin{axis}[%
    hide axis, xmin=0,xmax=0.001,ymin=0,ymax=0.001,
    legend style={draw=white!15!white,
    line width = 1pt,
    legend cell align=left,
    legend columns=2,
    /tikz/every even column/.append style={column sep=0.25cm},
    font = \small
    },
    ]
    \addlegendimage{Seaborn1}        \addlegendentry{GazeTR~\cite{cheng2022gaze}};
    \addlegendimage{Seaborn2}        \addlegendentry{L2CS~\cite{abdelrahman2023l2cs}};
    \addlegendimage{Seaborn3}
    \addlegendentry{3DGazeNet~\cite{ververas20243dgazenet}}
    \addlegendimage{Seaborn4}
    \addlegendentry{Gaze3D~\cite{vuillecard20253dgaze}}
    \end{axis}
    \end{tikzpicture}
    \vspace{-5cm}

    \begin{center}
    \includegraphics[width=0.85\linewidth]{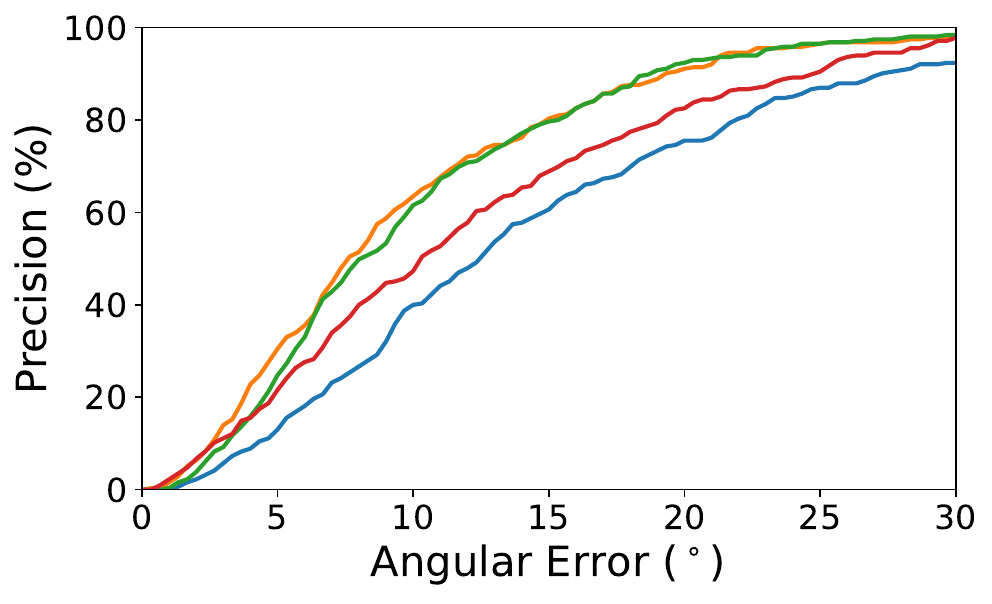} \includegraphics[width=0.9\linewidth]{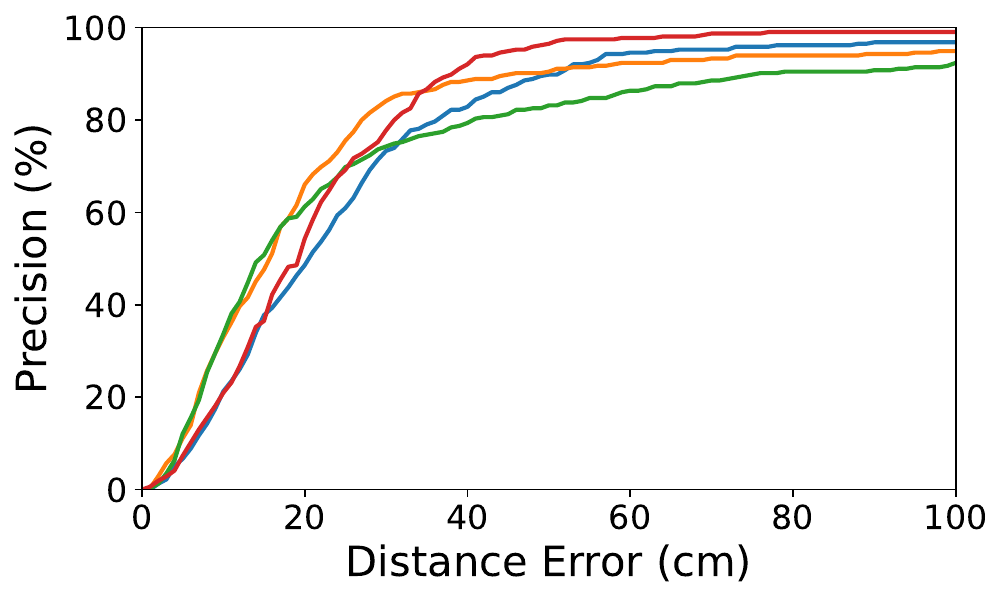}
    \end{center}

    \caption{The cumulative distribution of the angular (top) and the distance (bottom) errors. The plots show the precision achieved (y-axis) given an error threshold (x-axis).}
    \label{fig:cum_error}
\end{figure}

\section{Evaluation Dataset}

To evaluate the gaze estimation methods we have collected a dataset with human participants. We recruited six participants, five men and one woman in the age range 20--25. The participants were provided with information about the experiments and signed informed consent forms in accordance with a protocol approved by the Ethical Committee of the Faculty of Mathematics, Physics and Informatics, Comenius University Bratislava, with the number EKFMFI-23-02.

The participants were instructed to look at numbered white squares of the grid shown on the display in the shared workspace. The individual white squares are marked with numbers from 1 to 20 (see Fig.~\ref{fig:setup}). Each participant was instructed to look at a specific target square while the administrator took images with the robot's cameras. The participants were asked to repeat this from different distances and positions one to four times (2.6 times on average) for each square. For squares 11--20, the participants were asked to wear glasses. In total our evaluation dataset contains 315 images from each camera.


\begin{figure}[htbp]
    \centering
    \begin{tabular}{c}
        {\small GazeTR~\cite{cheng2022gaze}} \\
        \includegraphics[width=0.85\linewidth]{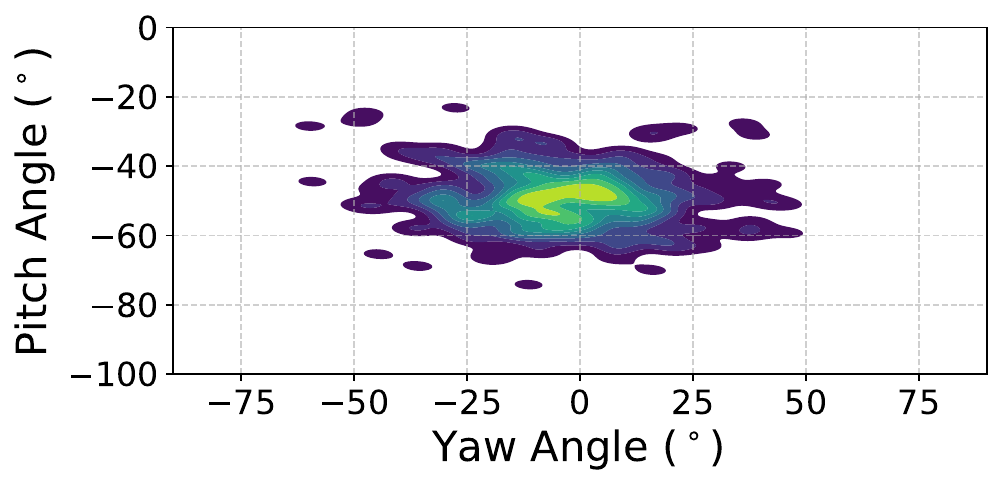} \\[-1pt]
        
        {\small L2CS~\cite{abdelrahman2023l2cs}} \\
        \includegraphics[width=0.85\linewidth]{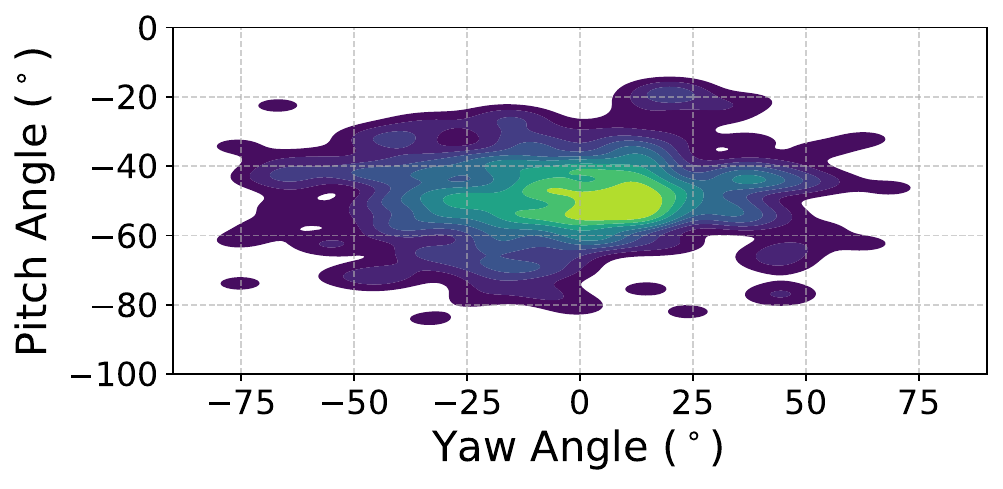} \\[-1pt]
        
        {\small 3DGazeNet~\cite{ververas20243dgazenet}} \\
        \includegraphics[width=0.85\linewidth]{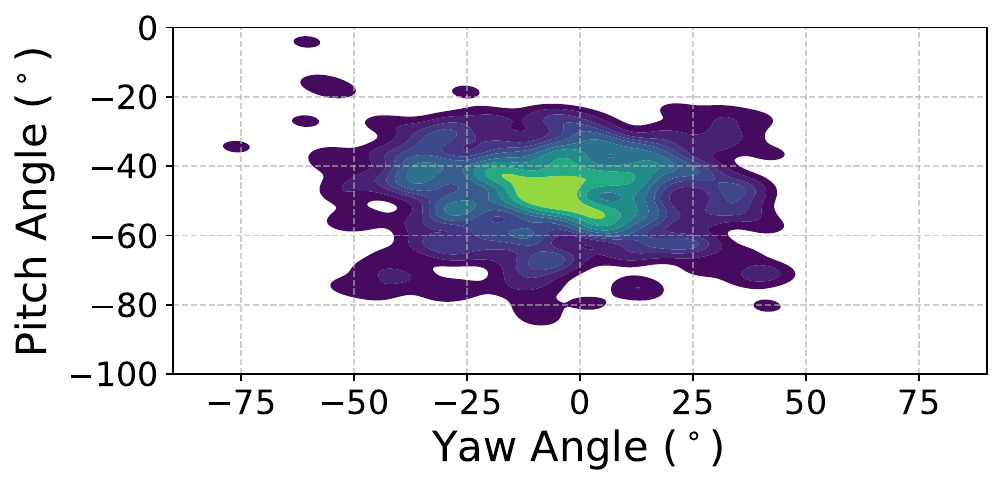} \\[-1pt]
        
        {\small Gaze3D~\cite{vuillecard20253dgaze}} \\
        \includegraphics[width=0.85\linewidth]{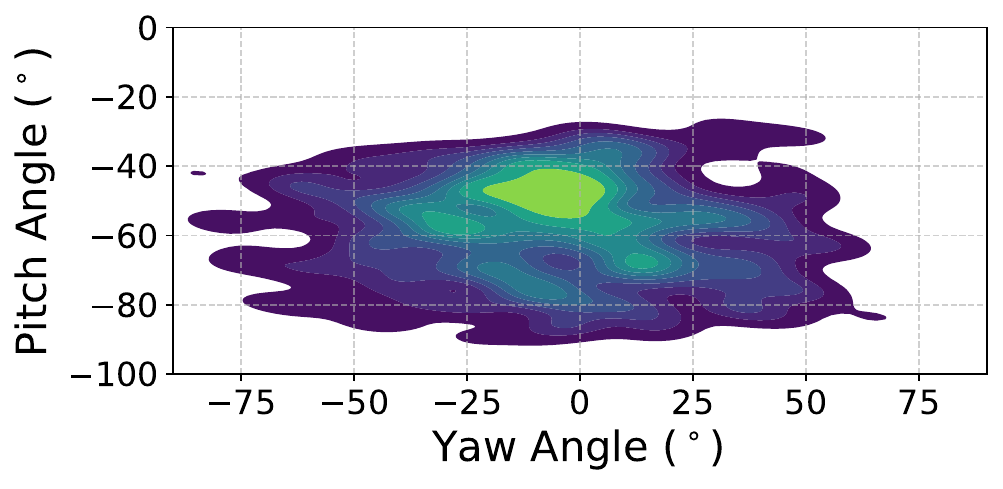} \\[-1pt]
        
        {\small Ground Truth} \\
        \includegraphics[width=0.85\linewidth]{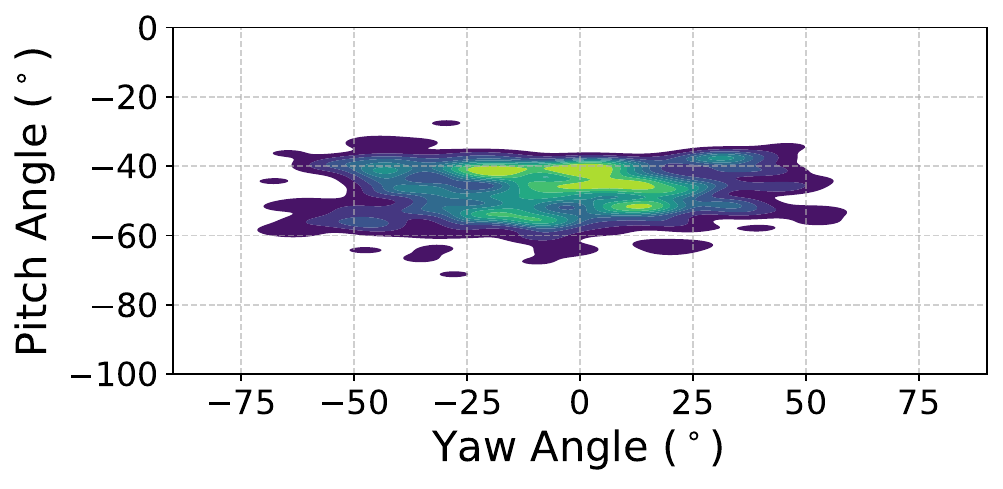}
    \end{tabular}
    \caption{The distributions of the yaw and pitch angles for the ground truth and predicted gaze. For ground truth, we consider the direction from the head position estimated using RetinaFace~\cite{deng2020retinaface} to the center of the target square. For visualization, we allow the pitch angles to go below $-90^\circ$.}
    \label{fig:distributions}
\end{figure}

\subsection{Evaluation Metrics}

To evaluate the gaze accuracy, we use the angle between the ground truth gaze direction and the gaze estimated by the network. The ground truth gaze is calculated using the estimated 3D position of the head and the 3D position of the center of the target square. We report the mean angular error of the predicted gaze directions.

We also calculate the error in terms of the estimated gaze point on the plane of the shared workspace (see Fig.~\ref{fig:setup}a). Based on the estimated gaze and the estimated 3D head position, we calculate the point where the gaze ray intersects the plane of the shared workspace (see Sec.~\ref{sec:method_gaze}) and calculate the distance from the center of the target square. Since the distance can grow without bound with increasing angular gaze error we report the median distance instead of the mean, which may be biased by a few very large error measurements. We also report the proportion of the points estimated within 10, 20, and 50 cm distance error. We denote these metrics as Precision@$X$cm.

\section{Evaluation}

\subsection{Evaluated Gaze Estimation Methods}
\label{sec:eval_nets}

In our evaluation we include four recent methods: GazeTR~\cite{cheng2022gaze}, L2CS~\cite{abdelrahman2023l2cs}, 3DGazeNet~\cite{ververas20243dgazenet}, and Gaze3D~\cite{vuillecard20253dgaze}. We use their official implementations for evaluation. All methods first crop out faces using dedicated face detection networks. Gaze3D uses YOLOv5 trained on the CrowdHuman dataset~\cite{shao2018crowdhuman}, L2CS and 3DGazeNet use RetinaFace~\cite{deng2020retinaface}. Since the official implementation for GazeTR does not provide a specific face detector, we use RetinaFace~\cite{deng2020retinaface}.

To determine the 3D position of the head we perform face detection on both the left and the right images from the robot's cameras. We triangulate the centers of the two boxes to obtain the 3D point. Since 3DGazeNet also estimates 2D coordinates of the eyes, we use the midpoints between the left and right eyes to calculate the 3D head position instead of relying on bounding box centers.

To evaluate GazeTR we use the GazeTR-Hybrid model trained on the ETH-XGaze dataset~\cite{zhang2020ethxgaze}. The L2CS, 3DGazeNet and Gaze3D models were trained using the Gaze360 dataset~\cite{kellnhofer2019gaze360}. Additionally, 3DGazeNet was trained using a semi-synthetic dataset ITWG-MV \cite{ververas20243dgazenet}. Gaze3D was also trained on the GazeFollow dataset~\cite{recasens2015they}.

L2CS, 3DGazeNet and Gaze3D return the gaze direction as the offset from the gaze directed directly at the camera center~\cite{kellnhofer2019gaze360}. To estimate the gaze direction in the coordinate system of the camera, we correct the estimated yaw and pitch by adding the yaw and pitch from the 3D head position to the camera center to the estimates provided by the networks.

\subsection{Results}

The results of our evaluation are shown in Table~\ref{tab:results}. We also show the cumulative distribution functions of both the angular and distance errors in Fig.~\ref{fig:cum_error}. L2CS and 3DGazeNet perform the best in terms of the angular error. We note that both methods reach mean angular error similar to the error they achieve on the Gaze360 dataset~\cite{kellnhofer2019gaze360} ($\sim 10^\circ$). In terms of the distance in the plane of the shared workspace, 3DGazeNet performs the best with a median distance of 14.57~cm. This suggests that localization of gaze within the shared workspace is limited in its accuracy.

In Fig.~\ref{fig:distributions}, we plot the distributions of the estimated gaze directions in comparison to the ground truth. The distributions show that GazeTR underestimates the magnitudes of the yaw angle while Gaze3D overestimates the negative pitch.

\subsection{Discussion}

Based on the presented results, it is clear that current off-the-shelf gaze estimation methods can be beneficial in HRI systems, but their limitations must be considered. When used in a scenario similar to ours, the gaze information can only provide an estimate with a resolution of tens of centimeters. This can inform the potential layout of the shared workspace, such that this accuracy is sufficient to discriminate between objects or relevant portions of the shared workspace. Conversely, gaze estimation networks could be used to provide only broader cues (e.g., whether the person is looking at the robot or at the shared workspace), which could still be useful. The inaccuracy could be accounted for and supplemented with information from other modalities in a multimodal perception system or by leveraging a broader context of the task. Performing multiple estimates using several images (video frames) could also lead to greater accuracy in aggregate or real-time information about the certainty of prediction accuracy.

\section{Conclusion}

In this paper, we evaluate the accuracy of using off-the-shelf gaze estimation methods in an HRI scenario involving a shared workspace of a robot and a human. In order to perform the evaluation, we have collected an annotated dataset using the robotic platform NICO. We have evaluated four recent methods using our dataset. We found that the methods perform within the expected accuracy in terms of the mean angular error. However, when the estimated gaze is used to determine a specific gaze point in the planar shared workspace, the error within the plane expressed in terms of metric distance remains relatively high, with a 15~cm median error for the best performing method. Based on our findings, we provide several recommendations on how to incorporate off-the-shelf gaze estimation methods in HRI systems.

Our work has several limitations. As future work, our dataset could be extended to include a greater variety of participants, additional robotic platforms, or different scenarios. Additionally, the evaluation could be strengthened by also considering videos providing additional temporal context.



\section*{Acknowledgment}

This work was funded by the EU NextGenerationEU through the Recovery and Resilience Plan for Slovakia under Project No.09I02-03-V01-00029. The work presented in this paper was carried out in the framework of the TERAIS project, a Horizon-Widera-2021 program of the European Union under the Grant agreement number 101079338. This work was also supported by the Slovak Grant Agency for Science (VEGA), project no.~1/0373/23.

\bibliographystyle{IEEEtran}
\bibliography{main}

\end{document}